\documentclass[letterpaper]{article} % DO NOT CHANGE THIS
\usepackage{aaai25}  % DO NOT CHANGE THIS
\usepackage{times}  % DO NOT CHANGE THIS
\usepackage{helvet}  % DO NOT CHANGE THIS
\usepackage{courier}  % DO NOT CHANGE THIS
\usepackage[hyphens]{url}  % DO NOT CHANGE THIS
\usepackage{graphicx} % DO NOT CHANGE THIS
\usepackage{natbib}  % DO NOT CHANGE THIS AND DO NOT ADD ANY OPTIONS TO IT
\usepackage{caption} % DO NOT CHANGE THIS AND DO NOT ADD ANY OPTIONS TO IT
\usepackage{algorithm}
\usepackage{algorithmic}

\usepackage{newfloat}
\usepackage{listings}
\DeclareCaptionStyle{ruled}{labelfont=normalfont,labelsep=colon,strut=off} % DO NOT CHANGE THIS
\floatstyle{ruled}
\newfloat{listing}{tb}{lst}{}
\floatname{listing}{Listing}
\usepackage{booktabs}
\usepackage{xcolor} 
\usepackage{multirow}
\usepackage[algo2e,ruled,vlined]{algorithm2e}
\usepackage{array}
\usepackage{amsmath}
\usepackage{amsfonts}
\usepackage{subcaption}
\usepackage{pifont}% http://ctan.org/pkg/pifont
\usepackage{xspace}
\usepackage{tikz}
\usepackage{fancybox}

\newcommand*\squared[1]{\tikz[baseline=(char.base)]{
    \node[draw, rectangle, rounded corners=2pt, inner sep=2pt] (char) {#1};}}

\newcommand{\pname}[1]{\textcolor{black}{DAVI}}

\definecolor{deeppink}{rgb}{1.0, 0.08, 0.58}
\title{Generalizable Disaster Damage Assessment \\
via Change Detection with Vision Foundation Model}
\author{
    Kyeongjin Ahn\textsuperscript{\rm 1},
    Sungwon Han\textsuperscript{\rm 1},
    Sungwon Park\textsuperscript{\rm 1}, \\
    Jihee Kim\textsuperscript{\rm 2,\rm 1,\rm 3},
    Sangyoon Park\textsuperscript{\rm 4},
    Meeyoung Cha\textsuperscript{\rm 5,\rm 1}\thanks{The corresponding author}
}
\affiliations{
    \textsuperscript{\rm 1}School of Computing, Korea Advanced Institute of Science and Technology (KAIST), Daejeon, South Korea\\

    \textsuperscript{\rm 2}College of Business, Korea Advanced Institute of Science and Technology (KAIST), Daejeon, South Korea

    \textsuperscript{\rm 3}Graduate School of Data Science, Korea Advanced Institute of Science and Technology (KAIST), Daejeon, South Korea

    \textsuperscript{\rm 4}Division of Social Science, Hong Kong University of Science and Technology (HKUST), Kowloon, Hong Kong

    \textsuperscript{\rm 5}Data Science for Humanity, Max Planck Institute for Security and Privacy (MPI-SP), Bochum, Germany
    \vspace{2mm}
}

\usepackage{bibentry}
\begin{document}

\maketitle

%     \link{Code}{https://github.com/kyeongjin0110/DAVI}
%     \link{Datasets}{https://xview2.org/}
%     \link{Extended version}{https://github.com/kyeongjin0110/DAVI}

% \input{sec/0_abstract}   
% \input{sec/1_introduction}
% \input{sec/2_related}    
% \input{sec/3_method}
% \input{sec/4_experiment}    
% \input{sec/5_conclusion}

\begin{abstract}
The increasing frequency and intensity of natural disasters call for rapid and accurate damage assessment. 
In response, disaster benchmark datasets from high-resolution satellite imagery have been constructed to develop methods for detecting damaged areas.
However, these methods face significant challenges when applied to previously unseen regions due to the limited geographical and disaster-type diversity in the existing datasets.
We introduce \pname{} ({D}isaster {A}ssessment with {VI}sion foundation model), a novel approach that addresses domain disparities and detects structural damage at the building level without requiring ground-truth labels for target regions.
\pname{} combines task-specific knowledge from a model trained on source regions with task-agnostic knowledge from an image segmentation model to generate pseudo labels indicating potential damage in target regions.
It then utilizes a two-stage refinement process, which operate at both pixel and image levels, to accurately identify changes in disaster-affected areas.
Our evaluation, including a case study on the 2023 Türkiye earthquake, demonstrates that our model achieves exceptional performance across diverse terrains (e.g., North America, Asia, and the Middle East) and disaster types (e.g., wildfires, hurricanes, and tsunamis).
This confirms its robustness in disaster assessment without dependence on ground-truth labels and highlights its practical applicability.
% \link{https://github.com/kyeongjin0110/DAVI}
% Code and appendix are available at \href{https://github.com/kyeongjin0110/DAVI}.
\looseness=-1
\end{abstract}

\section{Introduction}
\label{sec: intro}

\begin{figure}[h!]
% \vspace{4mm}
    \centering
    \includegraphics[width=0.9\columnwidth]{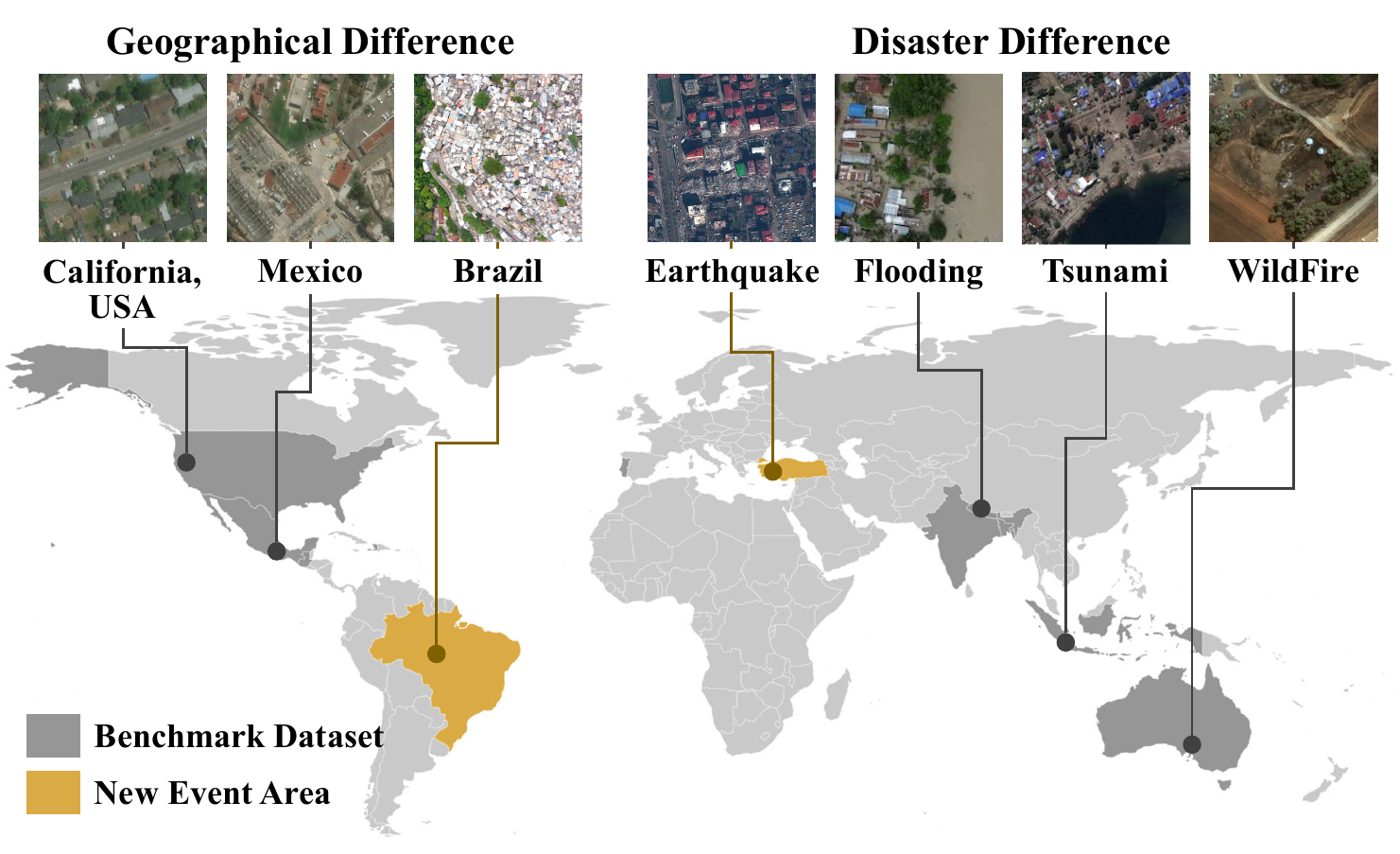}
    % \cutcaptionup
    % \vspace*{-2mm}
    \caption{
    Collaborative efforts are underway to collect datasets on natural disasters, yet no single benchmark dataset covers the entire world. The yellow areas represent regions with recent disasters not included in the dataset. \looseness=-1
    }
    \label{fig: benchmark_problem}
\vspace{-3mm}
\end{figure}

Natural disasters such as wildfires, hurricanes, floods, and earthquakes are becoming increasingly prevalent and severe due to climate change~\cite{change2022intergovernmental}.
These catastrophic events significantly impact human lives and cause extensive economic losses, necessitating prompt responses in rescue operations and financial assistance.
With limited resources available for life-saving relief and infrastructure recovery, assessing disaster damage at a fine-grained infrastructure-level becomes critical for effective resource distribution~\cite{fiedrich2000optimized, field2012managing}. 
Among the various efforts, multi-temporal remote sensing images have emerged as a valuable resource for disaster damage assessment, which can cover broad geographic regions at a lower cost relative to traditional ground surveys~\cite{weber2020building, wu2021building, gupta2021rescuenet}. \looseness=-1

Conventional methods rely on \textit{change detection approaches} that compare visual differences in regional images taken before and after a disaster.
Methods such as image differentiation~\cite{st2014subsense}, image ratio~\cite{skifstad1989illumination}, principal component analysis~\cite{kuncheva2013pca}, and change vector analysis~\cite{bruzzone2000automatic} are commonly applied to these problems.
Recent works detect changes using deep features through thresholding techniques~\cite{saha2019unsupervised}, specialized convolution networks~\cite{zhao2017discriminative}, and contrastive learning~\cite{wu2021unsupervised}. \looseness=-1

However, these techniques face significant challenges in real-world damage assessment due to substantial discrepancies between available training data (source data) and test data (target data).
Figure~\ref{fig: benchmark_problem} illustrates this issue, showing that the regions in the extensive disaster benchmark dataset, xBD~\cite{gupta2019creating}, span 45,000 square kilometers of annotated high-resolution satellite images of natural disasters, yet cover only a small part of the world. 
The limited coverage of these benchmark datasets makes it difficult to analyze features of previously unseen regions in terms of visual style. 
Additionally, the visual features of damages vary by disaster type; for instance, wildfires can destroy entire buildings, hurricanes may damage only roofs, and floods may engulf structures without causing visible damage. 
Therefore, \textit{domain adaptation} in images (i.e., shifts in visual styles) is required to handle diverse geographical landscapes, while \textit{adjustment} in the damage types (i.e., shift in the decision boundary) must be supported. \looseness=-1

We present \pname{} ({D}isaster {A}ssessment with {VI}sion foundation model), designed to bridge domain disparities and detect structural damages at the building level in an unsupervised fashion.
This is achieved through a test-time adaptation technique, which starts with a source model that is trained on independent source data and gradually updates the model to be used on target data without requiring target labels using pseudo labels~\cite{liang2020we, wang2022continual, niu2022efficient, chen2022self}.
We minimize the entropy of pseudo labels to reduce uncertainty arising from domain shifts.
Our approach also recognizes decision boundary shifts by employing an existing image segmentation model, SAM (Segment Anything Model)~\cite{kirillov2023segment}, and integrating the pseudo labels with general visual features. 
Image segmentation is especially useful for disaster assessment because it can generate semantic masks for various objects, including buildings, roads, vegetation, and water bodies.
We demonstrate that this foundation model is also effective for investigating structural damage in disasters.
\looseness=-1

\pname{} refines pseudo labels from the source model through a two-stage process. First, the pixel-level adjustment uses inconsistent change maps from temporally paired images and their augmented ones, then the image-level adjustment leverages coarse-grained damage presence information.
These pseudo labels provide supervisory guidance to align model predictions in the absence of ground-truth labels.
We evaluate the effectiveness of \pname{} using real-world events from diverse regions and disaster types within the xBD benchmark dataset. The results show that \pname{} successfully identifies structural damage in previously unseen areas, outperforming unsupervised change detection and domain adaptation baselines.
Additionally, we validate its practical applicability with recent disaster samples not included in the benchmark dataset. \looseness=-1

Disaster damage in uncharted territories exhibits immense diversity in visual and inherent characteristics.
To effectively guide metrics until on-the-ground assessments, a detection method capable of handling such varied content is essential.
\pname{}, built on a vision foundation, represents a significant step forward in advancing this capability.
\looseness=-1

\section{Related Work}
\label{sec: related}

\subsection{Transfer Learning in Change Detection}
Transfer learning methods focus on applying a pre-trained model from the source data to the target data.
Given the high cost of acquiring ground-truth labels, especially in damage assessment, studies have utilized unsupervised methods.
For example, several approaches use pre-trained weights on pretext tasks like segmentation or classification to detect changes in temporally paired images~\cite{hou2017change, de2019unsupervised, zhang2020feature, wu2021unsupervised, leenstra2021self}.
However, transfer learning algorithms often assume that the training and test sets are independent and identically distributed ($i.i.d.$), which degrades performance when domain discrepancies exist~\cite{patricia2014learning}.
% An advanced version of transfer learning is the domain adaptation method, which aims to reduce domain discrepancies between datasets.
Domain adaptation techniques, an advanced version of transfer learning, partially address these discrepancies but assume that if pre-training and fine-tuning tasks share the same class label, the class's semantic characteristics remain consistent, even if the visual styles differ~\cite{wang2020tent, liang2020we, wang2022continual}.
For example, an apple illustration and a real-life photo have different visual styles but share semantic traits that make them recognizable as apples.
In our problem setting, however, the semantic differences caused by different types of disasters make domain adaptation methods difficult to apply, requiring a decision boundary shift.
\looseness=-1

\subsection{Vision Foundation Model} 
Deep neural networks and self-supervised learning has significantly improved vision foundation models (VFMs) on large-scale vision data~\cite{luddecke2022image, wang2023seggpt, kirillov2023segment, wang2023images, zou2024segment, wang2024visionllm}. 
Examples of VFMs for image segmentation include, but are not limited to, CLIPSeg~\cite{luddecke2022image}, SegGPT~\cite{wang2023seggpt}, SAM~\cite{kirillov2023segment}, and SEEM~\cite{zou2024segment}.
These models use visual inputs and prompts — such as text, boxes, points, or masks — to specify target image segmentation and show remarkable adaptability to other domains, such as remote sensing~\cite{osco2023segment, chen2024rsprompter}, tracking~\cite{yang2023track, rajivc2023segment}, robotics~\cite{yang2023pave}, and medical~\cite{lei2023medlsam, gong20233dsam, shaharabany2023autosam, wu2023medical, ma2024segment}, underscoring its effective capability to recognize various visual content.
We employ SAM in our change detection method as an assistant due to its simplicity and ease of use. 
SAM uses an image encoder, a prompt encoder, and a mask decoder. 
It combines image and prompt embeddings through each encoder to guide object localization, and uses the mask decoder to generate masks from these embeddings for self-annotation.
\looseness=-1

\begin{figure*}[t!]
    \centering
    \includegraphics[width=0.96\textwidth]{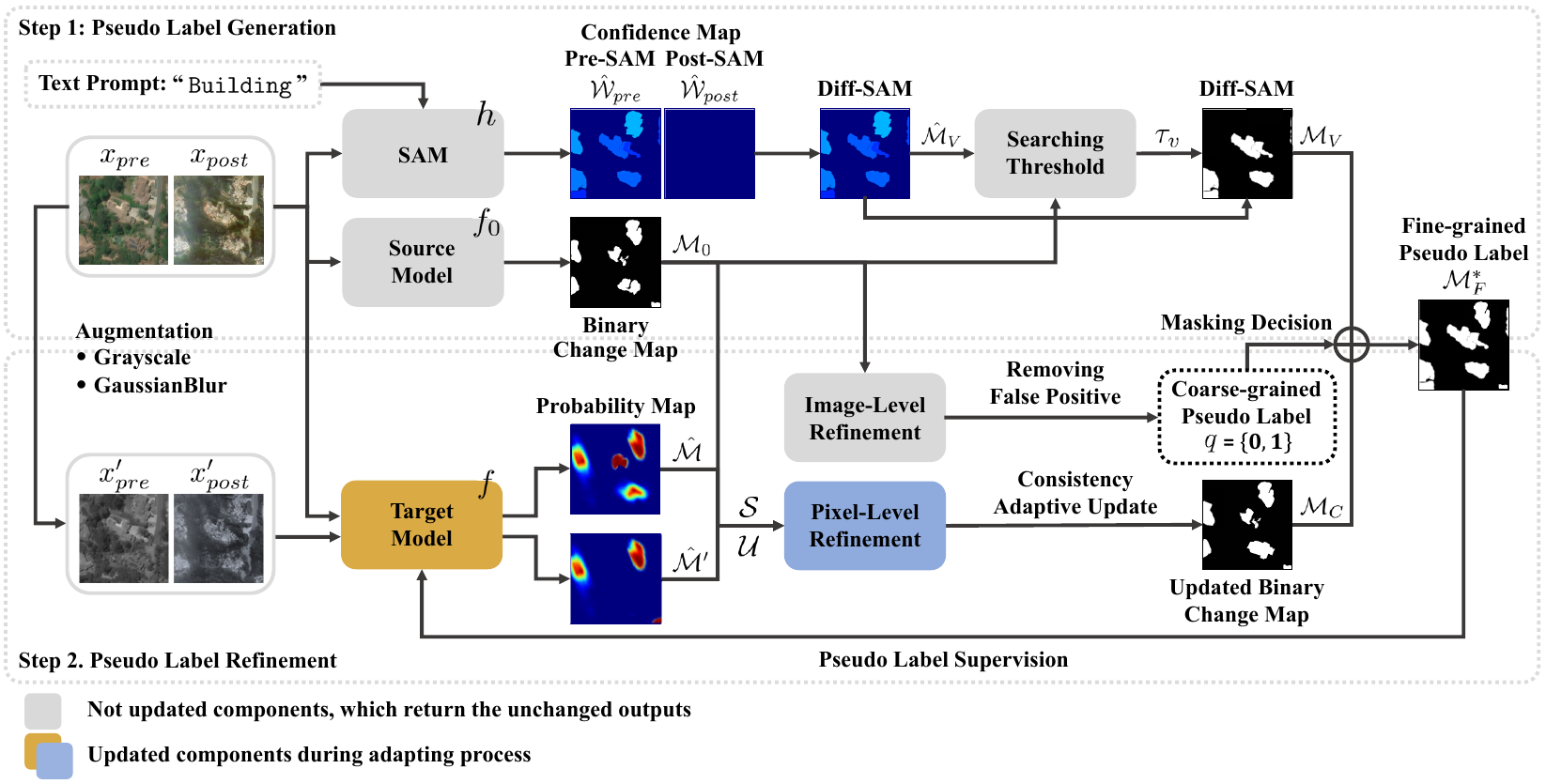}
    % \cutcaptionup
    \caption{
    Model architecture. This figure illustrates how pre- and post-disaster images and their augmented versions are processed. The system has two phases: (Step 1) pseudo label generation, which leverages the source model and image segmentation foundation model (SAM) with a text prompt for generalizability across various disaster scenarios, and (Step 2) pseudo label refinement, which involves pixel- and image-level refinement to reduce noise in the fine-grained pseudo label. The final output is a fine-grained pseudo label indicating potential damage. \looseness=-1
    }
    \label{fig: main_architecture}
\end{figure*}

\subsection{Change Detection by Pseudo Labeling}
When ground-truth labels are not available, one common strategy to provide supervisory signals is to introduce pseudo labels for training~\cite{lee2013pseudo, zou2020pseudoseg, hu2023binary, kim2022disaster}.
This pseudo labeling has been used in change detection problems as well to achieve higher accuracy. Examples include Pix4Cap~\cite{liu2023pixel}, which uses pseudo labels extracted from a pre-trained model to improve the model's ability to change captioning, and PPL~\cite{li2024progressive}, which generates pseudo labels through uncertainty estimation and progressive refinement on hyperspectral images.
These change detection methods using pseudo labels rely on self-updating through their own predictions, making implementation easy but accumulating errors over time. 
In this work, we utilize pseudo-labeling for learning together with an external foundation model and refinement process to reduce error propagation. \looseness=-1

\section{Methodology}
\label{sec: method}

\subsection{Problem Formulation}
\label{sec: method_problem_definition}
We present an efficient method for disaster assessment, \pname{}, which compares pre- and post-disaster images to detect structural damage (e.g., building, infrastructure level).
\pname{} runs in a test-time adaptation setting, using change detection models trained on unveiled disaster-related data (i.e., source data), adapted to target data. Below is the problem formulation of our setting with an overview of \pname{}. \looseness=-1

Let us define the source data $D_{s}$ as unveiled data from the source domain and the target data $D_{t}$ as veiled data from the target domain.
The source data $D_{s}$ contain temporally paired images and their corresponding ground truth labels, indicated as
$\left \{ \bar{x}_{pre}, \bar{x}_{post}, y \right \} \in D_{s}$, where $\bar{x}_{pre}$  $\in \mathbb{R}^{H\times W \times 3}$ represents a pre-disaster image, $\bar{x}_{post}$ $\in \mathbb{R}^{H\times W \times 3}$ represents a post-disaster image and $y$ $\in \{0,1\}^{H\times W}$ represents a binary change map, respectively.
The target data $D_{t}$ consist of temporally paired images, indicated as
$\left \{ x_{pre}, x_{post}\right \} \in D_{t}$ without access to ground-truth labels. Here, we ensure that there is no overlap between the source and target data. 
The model trained with the source data (i.e., source model) is denoted as $f_{0}$.
The target model, denoted as $f$, starts with the same weights as the source model $f_0$, and is then updated to adapt to the target data during training.
Our goal is to develop a domain-shifting strategy that relies solely on the source model and unlabeled target data, ultimately achieving high-accuracy predictions for changed areas within the target data using the target model $f$. \looseness=-1

Figure~\ref{fig: main_architecture} illustrates the model architecture.
We design pseudo labels $\mathcal{M}_{F}^{*}$ $\in \{0,1\}^{H\times W}$ that accommodate domain shifts across various types of disaster occurring globally.
These pseudo labels substitute for unavailable ground-truth labels in the target data. The intuition behind using pseudo labels is that they provide supervisory guidance, helping the model detect changes in the target data.
The generation phase (Step 1) leverages both the source model $f_{0}$ with task-specific knowledge based on the source domain and SAM $h$ with task-agnostic knowledge from extensive image corpus.
This is followed by a refinement process (Step 2) that improves the accuracy of the pseudo labels by removing noisy parts within them using consistency-based information and coarse-grained information.
Ultimately, in addition to supervised learning on the generated pseudo labels, we minimize the entropy of the predictions to learn domain-specific features for the target data.
Our objective of training the target model $f$ can be formulated as: \looseness=-1
\begin{equation}
\label{eq: entropy_loss}
H(\mathcal{\hat{M}}) = -\sum_{i}\sum_{j}p^{ij}\log{(p^{ij})}
\end{equation}
where $H(\mathcal{\hat{M}})$ refers to Shannon entropy loss function~\cite{shannon1948mathematical} frequently employed in one of the test-time adaptation approaches, which tries to connect errors arising from domain discrepancies and domain shifts~\cite{wang2020tent}.
Here, $\mathcal{\hat{M}}$ indicates the change maps computed by the target model $f$, and $p^{ij}$ denotes the probability value within $\mathcal{\hat{M}}$, where $(i, j)$ represents the coordinates of individual pixels.
Through this objective function, the target model $f$ eliminates generalization errors by minimizing the entropy of its predictions on the target data. \looseness=-1
\begin{equation}
\label{eq: total_loss}
\mathcal{L} = CE(\mathcal{M}_{F}^{*}, \mathcal{\hat{M}}) + \lambda H(\mathcal{\hat{M}})
\end{equation}
where $\mathcal{M}_{F}^{*}$ is generated fine-grained pseudo labels and $\lambda$ is an adjusting hyperparameter, which is set to 0.1 as default. \looseness=-1

\subsection{Step 1: Pseudo Label Generation}
\label{sec: method_generation}
We generate the fine-grained pseudo labels $\mathcal{M}_{F}$ that represent pixel-level binary change maps.
This information combines two types of pseudo labels: $\mathcal{M}_{0}$ from the source model $f_{0}$ and $\mathcal{M}_V$ from SAM, $h$.
This enhances model generalizability across various disaster scenarios by utilizing complementary knowledge.
Relying solely on pseudo labels from the source model can lead to inaccurate signals due to mismatches between source and target distributions, ultimately hindering the model's applicability.
If the model overfits the source data, it may exhibit undue uncertainty or certainty when applied to the target data.
This affects the interpretation of the model's predictions, potentially requiring a recalibration of confidence thresholds or adjustments to output probabilities.
Trained on billion-scale vision data, SAM can analyze visual features across various domains. By integrating this with the source model's knowledge, the combined approach adeptly manages decision boundary shifts, enhancing adaptation in diverse scenarios.
\looseness=-1
\smallskip

\noindent\textbf{Pseudo label generation with task-specific knowledge.} Given the paired images $x_{pre}$ and $x_{post}$, the source model $f_{0}$ predicts the probability $p_0^{ij}$ of the change for each pixel in the change maps $\mathcal{\hat{M}}_{0}$. (i.e., $\mathcal{\hat{M}}_{0}^{ij} = p_{0}^{ij}$)
\begin{equation}
\label{eq: change_map_from_source}
\mathcal{\hat{M}}_{0}^{ij} = f_{0}(x_{pre}^{ij},\ x_{post}^{ij})
\end{equation}
We convert the probability maps $\mathcal{\hat{M}}_{0}$ into the binary change maps $\mathcal{M}_{0}$ as follows: \looseness=-1
\begin{equation}
\label{eq: binary_cd_from_source}
  \mathcal{M}_{0}^{ij}  =
  \begin{cases}
    1 & \text{if}\ p_{0}^{ij} \geq 0.5 \\
    0 & \text{otherwise}
  \end{cases},  \  \ m_{0}^{ij} \in \mathcal{M}_{0}
\end{equation}

\noindent\textbf{Pseudo label generation with task-agnostic knowledge.}  
Next, our method segments objects of interest (e.g., buildings, roads, vegetation, and water bodies) based on the vision foundation model like SAM to adjust the source model's pseudo label.
It segments specific objects using a text prompt and assigns a confidence score to each object: \looseness=-1
\begin{equation}
\label{eq: confidence_map_from_sam}
\mathcal{\hat{W}}=h(x;prompt), \  \ w^{ij} \in \mathcal{\hat{W}} \ \ \text{and} \ \  w^{ij} \in \left [0, 1  \right ].
\end{equation}
To obtain semantic changes between the paired images, we subtract two segmented maps with confidence scores — specifically, Pre-SAM and Post-SAM.
When using the text prompt ``\texttt{Building}", SAM has high confidence for identifying building areas in pre-disaster images, while it would have decreased confidence for damaged building areas in post-disaster images because it lacks the essential features for recognizing buildings.
Therefore, we leverage the confidence difference between pre-disaster and post-disaster images as an indicator of building destruction, utilizing SAM $h$ with the text prompt ``\texttt{Building}": \looseness=-1
\begin{equation}
\label{eq: confidence_difference}
\mathcal{\hat{M}}_{V}^{ij} = \max_{ij} (0,\ w_{pre}^{ij} - w_{post}^{ij}),  \  \ d_{v}^{ij} \in \mathcal{\hat{M}}_{V}
\end{equation}
\noindent where $\mathcal{\hat{M}}_{V}$ represents the confidence difference maps from SAM, referred to as Diff-SAM. We clip the confidence difference maps to be greater than zero to focus on the changes in destruction rather than construction.
Then, we transform the confidence difference maps $\mathcal{\hat{M}}_{V}$ into the binary change maps $\mathcal{M}_{V}$ as follows: \looseness=-1
\begin{equation}
\label{eq: binary_change_map_from_confidence_difference}
  \mathcal{M}_{V}^{ij} =
  \begin{cases}
    1 & \text{if}\ d_{v}^{ij} \geq \tau_{v} \\
    0 & \text{otherwise}
  \end{cases}
\end{equation}
\noindent where $\tau_{v}$ represents a threshold determined through optimal threshold searching. For simplicity, we also hereafter refer to $\mathcal{M}_{V}$ as Diff-SAM.
Finally, we combine two binary change maps, using a max function to generate the fine-grained pseudo labels $\mathcal{\hat{M}}_{F}$:
\begin{equation}
\label{eq: combine_binary_change_map}
\mathcal{\hat{M}}_{F}^{ij} = \max_{ij} (\mathcal{M}_{0}^{ij},\ \mathcal{M}_{V}^{ij})
\end{equation}
This combination enables the target model to focus on damaged patterns that the source model could not capture, thereby enhancing recall for affected areas. \looseness=-1
% This combining, the target model focus on considering the unlearned damaged pattern that the source model could not capture.
\smallskip

\noindent\textbf{Searching Optimal Threshold.} 
The threshold $\tau_{v}$ sets the confidence level at which segmented buildings are identified as damaged.
The appearance of damaged buildings can differ depending on factors such as regional characteristics and the type of disaster, meaning the optimal threshold can also vary accordingly.
We determine the optimal threshold by exploring various candidate values that maximize the F1-score between $\mathcal{M}_{0}$ and $\mathcal{M}_{V}$ on the target data.
This approach is based on the intuition that finding the precise threshold requires disaster-specific knowledge, which can be aided by the source model, even if it pertains to a different domain. \looseness=-1

\subsection{Step 2: Pseudo Label Refinement}
\label{sec: method_refinement}
We refine the fine-grained pseudo labels through in two processes:
\textbf{Consistency Adaptive Update} and \textbf{Coarse-grained Pseudo Labels} to minimize noise in these labels.
Despite utilizing knowledge from the source model and SAM, the noise remains in the fine-grained pseudo labels.
This issue can be attributed to the distortion of knowledge adapted from the source model to the target distribution due to the domain discrepancy, or to SAM's insufficient capability to adequately consider features associated with changes in the target data.
We first focus on pixel-level adjustments, targeting areas identified as changed within the fine-grained pseudo labels, guided by confidence-based consistency from the data augmentation strategy.
We then engage in image-level adjustments to further enhance noise reduction effectively. 
This approach leverages on broadly unchanged information between the paired images, from the coarse-grained pseudo labels. These labels are used to mask the generated fine-grained pseudo labels, rendering them as all zeros. \looseness=-1 \smallskip

\noindent\textbf{Consistency Adaptive Update.}  
Unsupervised or semi-supervised learning can lead to overconfident and poorly calibrated predictions~\cite{park2021improving, gao2020consistency}. 
Consequently, pseudo-labeling-based on confidence scores can be suboptimal. 
To address this, we propose using inconsistencies arising from data augmentation variations to assess uncertainty for pixel-level refinement. 
These inconsistencies highlight the model's uncertainty in its decisions.
Provided with the paired images $x_{pre}$ and $x_{post}$, along with their augmented counterparts $x'_{pre}$ and $x'_{post}$, we adaptively update the predicted binary change maps $\mathcal{M}_{0}$ from the source model $f_{0}$ using the confidence metrics obtained from these original and augmented paired images.
Here, the target model $f$ predicts the probability $p^{ij}$ of the change for each pixel in the change maps $\mathcal{\hat{M}}$ as same as the source model $f_{0}$.
\begin{equation}
\label{eq: change_map_from_target}
\mathcal{\hat{M}}^{ij} = f(x_{pre}^{ij},\ x_{post}^{ij})
\end{equation} \looseness=-1
Similarly, we obtain probability maps from another augmentation, denoted as $\mathcal{\hat{M}}' = f(x'_{pre}, x'_{post})$.
Using these maps, we then compute mean $\hat{\mathcal{U}}$ and standard deviation $\mathcal{S}$ from $\mathcal{\hat{M}}$ and $\mathcal{\hat{M}}'$ to quantify confidence based on pixel-wise variability.
As in Eq. (\ref{eq: binary_cd_from_source}), the averaged probability maps $\hat{\mathcal{U}}$ are converted into the binary change maps $\mathcal{U}$. \looseness=-1

Our refinement operates under a strategy using $\mu^{ij} \in \mathcal{U}$ and $\sigma^{ij} \in \mathcal{S}$. 
If variation $\sigma^{ij}$ is marginal (i.e., it has low uncertainty), we use the value $m_{0}^{ij}$ in the binary change map $\mathcal{M}_{0}$ from the source model $f_{0}$ (see Eq. (\ref{eq: binary_cd_from_source})) to its corresponding value $\mu^{ij}$ in the binary change map $\mathcal{U}$ from the target model $f$.
Conversely, if variation $\sigma^{ij}$ is high (i.e., it has high uncertainty), we use the original value $m_{0}^{ij}$ of the binary change map $\mathcal{M}_{0}$.
In each iteration, the updated binary change maps $\mathcal{M}_{C}$ are generated from $\mathcal{M}{0}$ in Eq. (\ref{eq: combine_binary_change_map}), guided by the evolving $\mathcal{U}$.
This process ensures they progressively reflect the target knowledge as follows: \looseness=-1
\begin{equation}
\label{eq: updated_binary_map_from_source}
    \mathcal{M}_{C}^{ij} = \left\{
    \begin{array}{l@{}l}
    \mu^{ij} & \quad \text{if} \ \sigma^{ij}  < \tau_{r} \ \ \text{and} \ \ \mu^{ij} = 1 \\
    m_{0}^{ij} & \quad \text{otherwise}
    \end{array}
\right.
\end{equation}
\noindent where $\tau_{r}$ is the threshold for pixel update decisions, set to $0.001$. Using these updated maps, we can derive the refined fine-grained pseudo labels. \looseness=-1
\begin{equation}
\label{eq: update_binary_change_map}
\mathcal{M}_{F}^{ij} = \max_{ij} (\mathcal{M}_{C}^{ij},\ \mathcal{M}_{V}^{ij})
\end{equation} 

\noindent\textbf{Coarse-grained Pseudo Labels.}
We employ the coarse-grained pseudo labels generated from the binary change maps $\mathcal{M}_{0}$ of the source model $f_{0}$ for the image-level refinement.
These labels provide image-level information indicating the absence (0) or presence (1) of changes in the paired images. 
We use this information to guide the model more effectively by masking the generated fine-grained pseudo labels $\mathcal{M}_{F}$ in instances where the binary change maps $\mathcal{M}_{0}$ show no changes.
This refinement process helps suppress false positives (FP) in the fine-grained pseudo labels rather than relying on the unmasked fine-grained pseudo labels.
We can formulate the coarse-grained pseudo labels $q$ as follows: \looseness=-1
\begin{equation}
\label{eq: coarse_label}
    q = \left\{
    \begin{array}{l@{}l}
    1 & \quad \text{if} \ \sum_{i, j} {m_0^{ij}} > 0 \\
    0 & \quad \text{otherwise}
    \end{array}
\right.
\end{equation}
\noindent Using these labels, we mask the fine-grained pseudo labels $\mathcal{\hat{M}}_{F}$.
\begin{equation}
\label{eq: fine_label}
\mathcal{M}_{F}^{*} = q \cdot \mathcal{M}_{F}, \ \ m_{f}^{ij} \in \mathcal{M}_{F}^{*}
\end{equation}
\noindent where $\mathcal{M}_{F}^{*}$ is the final fine-grained pseudo labels after the image-level refinement.
Using these fine-grained pseudo labels, we proceed to train the target model $f$ as follows: \looseness=-1
\begin{equation}
\label{eq: cd_loss}
CE(\mathcal{M}_{F}^{*}, \mathcal{\hat{M}}) = -\sum_{i}\sum_{j}m_{f}^{ij}\log{(p^{ij})}
\end{equation}

\section{Experiments}
\label{sec: experiment}

\subsection{Experimental Setup}
\label{subsec: experimental_setup}
\noindent\textbf{Datasets.}
Our evaluation utilizes the well-known change detection benchmark dataset, xBD. This dataset is widely used for change detection problems and building damage assessment, encompassing different regions affected by 11 various disasters, including wildfires, hurricanes, floods, and earthquakes. For training the source model, we specifically selected data from wildfires in Woolsey, California, USA, and floods in Monsoon, Nepal.
Our analysis targets wildfires in Santa Rosa, California, USA, hurricanes in Texas, USA, and tsunamis in Sulawesi, Indonesia.
To ensure the model's broader applicability for disaster assessment across diverse terrains, we selected source and target data that differ in both geographical locations and disaster types.
%Details about each volume are in \todo{Appendix}. 
\looseness=-1
\smallskip

\noindent\textbf{Ground-truth Labels.} Due to the deviation in disaster level information across different types of disasters, we convert the four-category disaster level information (0: \textit{no damage}, 1: \textit{minor damage}, 2: \textit{major damage}, 3: \textit{destroyed}) into the binary information (0: \textit{no damage}, 1: \textit{damage}) at each pixel within ground-truth labels. \looseness=-1
\smallskip

\begin{table*}[t!]	
\centering
% \cutcaptionup
\resizebox{0.99\textwidth}{!}	
{	
    \begin{tabular}{@{}c|c|cccc|cccc|cccc@{}}
    \toprule
    \multirow{2}{*}{Methods} & \multirow{2}{*}{Backbones} & \multicolumn{4}{c|}{Wildfires (California, USA)} & \multicolumn{4}{c|}{Hurricanes (Texas, USA)} & \multicolumn{4}{c}{Tsunamis (Sulawesi, Indonesia)} \\ \cline{3-14} 
     &  & Precision & Recall & F1-score & Accuracy & Precision & Recall & F1-score & Accuracy & Precision & Recall & F1-score & Accuracy \\ \midrule
     
    \multirow{2}{*}{Oracle} & SNUNet-CD & 0.8476 & 0.7733 & 0.8087 & 0.9939 & 0.8075 & 0.6190 & 0.7008 & 0.9767 & 0.7402 & 0.5025 & 0.5986 & 0.9878 \\
     & BIT-CD & 0.8015 & 0.8272 & 0.8142 & 0.9938 & 0.8162 & 0.7075 & 0.7580 & 0.9801 & 0.7524 & 0.6436 & 0.6937 & 0.9897 \\ \midrule
     
    Source & SNUNet-CD & 0.8110 & 0.5119 & 0.6276 & 0.9899 & 0.0466 & 0.9338 & 0.0888 & 0.1542 & 0.4546 & 0.0272 & 0.0513 & 0.9818 \\ \midrule
    \multirow{2}{*}{Source\ding{61}} & SNUNet-CD & 0.6566 & 0.8109 & 0.7256 & 0.9899 & 0.6053 & 0.0414 & 0.0775 & 0.9565 & 0.5200 & 0.0191 & 0.0369 & 0.9819 \\
     & BIT-CD & 0.8227 & 0.5063 & 0.6268 & 0.9900 & 0.7317 & 0.0168 & 0.0328 & 0.9563 & 0.7046 & 0.0834 & 0.1491 & 0.9828 \\ \midrule
    
    CVA & - & 0.0343 & 0.4583 & 0.0638 & 0.7774 & 0.0527 & 0.2544 & 0.0873 & 0.7653 & 0.0304 & 0.3450 & 0.0559 & 0.7888 \\
    IRMAD & - & 0.0290 & 0.4885 & 0.0547 & 0.7090 & 0.0436 & 0.2688 & 0.0750 & 0.6948 & 0.0407 & 0.4293 & 0.0743 & 0.7908 \\
    PCAKmeans & - & 0.0322 & 0.5501 & 0.0609 & 0.7191 & 0.0306 & 0.2408 & 0.0543 & 0.6299 & 0.0197 & 0.3207 & 0.0371 & 0.6986 \\
    ISFA & - & 0.0336 & 0.4603 & 0.0626 & 0.7563 & 0.0471 & 0.2163 & 0.0773 & 0.7569 & 0.0384 & 0.4194 & 0.0703 & 0.7855 \\

    DCVA & CNN & 0.0320 & 0.5552 & 0.0605 & 0.7028 & 0.0430 & 0.2643 & 0.0740 & 0.6970 & 0.0023 & 0.8394 & 0.0045 & 0.6994 \\
    % KPCAMNet & CNN* &  &  &  &  &  &  &  &  &  &  &  &  \\
    SCCN & CNN & 0.0383 & 0.4821 & 0.0709 & 0.7772 & 0.0500 & 0.2317 & 0.0822 & 0.7551 & 0.0030 & 0.7444 & 0.0061 & 0.7853 \\ \midrule
    
    TENT & SNUNet-CD & 0.8810 & 0.1864 & 0.3077 & 0.9861 & 0.5994 & 0.0053 & 0.0104 & 0.9559 & 0.4449 & 0.0180 & 0.0346 & 0.9818 \\
    SHOT & SNUNet-CD & 0.6824 & 0.8025 & 0.7376 & 0.9906 & 0.6057 & 0.0413 & 0.0774 & 0.9565 & 0.5218 & 0.0191 & 0.0369 & 0.9819 \\
    CoTTA & SNUNet-CD & 0.4432 & 0.2772 & 0.3209 & 0.9897 & 0.3568 & 0.1006 & 0.1135 & 0.9562 & 0.2232 & 0.0355 & 0.0488 & 0.9817 \\  \midrule
    
    \multirow{2}{*}{\pname{}} & SNUNet-CD & 0.7206 & 0.7767 & \textbf{0.7476} & \textbf{0.9913} & 0.5211 & 0.3552 & 0.4225 & 0.9571 & 0.5640 & 0.3389 & \textbf{0.4234} & \textbf{0.9833} \\
     & BIT-CD & 0.7075 & 0.6715 & 0.6890 & 0.9900 & 0.5968 & 0.5925 & \textbf{0.5947} & \textbf{0.9643} & 0.3701 & 0.4187 & 0.3929 & 0.9766 \\ \bottomrule
     \multicolumn{14}{l}{ \ding{61} This is the fine-tuned source model that was trained using binary change maps from the pre-trained source model as fine-grained pseudo labels.} %\\
     %\multicolumn{10}{l}{ * We used the proposed architecture from the original paper.} \\
     
     \end{tabular}
}
\caption{
Performance of \pname{}, other unsupervised change detection, and domain adaptation baselines, tested across different geographic regions (North America and Asia) and disaster types (wildfires, hurricanes, and tsunamis). 
Top results are bolded. 
\looseness=-1
}	
\label{tab: main_cd}	
\end{table*}

\noindent\textbf{Baselines.}
Given that ground-truth labels in target regions are not accessible, we employ four traditional unsupervised change detection baselines — CVA~\cite{bruzzone2000automatic}, IRMAD~\cite{nielsen2007regularized}, PCAKmeans~\cite{celik2009unsupervised}, and ISFA~\cite{wu2013slow} — as well as two deep learning-based unsupervised change detection baselines, DCVA~\cite{saha2019unsupervised} and SCCN~\cite{zhao2017discriminative}.
We also adopt three relevant domain adaptation baselines — TENT~\cite{wang2020tent}, SHOT~\cite{liang2020we}, and CoTTA~\cite{wang2022continual} — and fine-tune them on the same backbone, following the same pre-training process.
\looseness=-1
\smallskip

\noindent\textbf{Implementation Details.}
Our experiments utilize SNUNet-CD~\cite{fang2021snunet} as a backbone for both the source model $f_{0}$ and the target model $f$, which predict the change maps $\mathcal{\hat{M}}_{0}$ and $\mathcal{\hat{M}}$, respectively. Additionally, BIT-CD~\cite{chen2021remote} is employed as another backbone for both models.
For all experiments, the batch size is $8$. On SNUNet-CD, the source and target model learning rates are $1e-3$ and $le-5$, respectively, while on BIT-CD, they are $1e-2$ and $le-4$, both with a weight decay of $0.01$.
A StepLR scheduler (step size $8$, gamma $0.5$) is used. All models are trained for $50$ epochs using the AdamW optimizer.
To regulate the degree of adaptation across experiments, $\lambda$ is set to $0.1$ in Eq. (\ref{eq: total_loss}). 
%For baselines, we used the proposed architecture from the original paper unless specified.
\looseness=-1 \smallskip

\noindent\textbf{Performance Evaluation}. 
Our proposed model is compared with various baselines, including traditional and deep learning based unsupervised methods, as well as domain adaptation techniques, using the following common assessment metrics: macro averages of precision, recall, F1-score, and accuracy. \looseness=-1 \smallskip

% \noindent\textbf{Others}. Further details are provided in  \todo{Appendix~\ref{subsec: appendix_hardware}}.

\subsection{Comparison with Baselines}
\label{sec: main_comparison}
Table~\ref{tab: main_cd} compares \pname{} with other relevant baselines.
Traditional methods (CVA, IRMAD, PCAKmeans, and ISFA) and deep learning-based methods (DCVA and SCCN) perform poorly overall compared to the proposed method due to their lack of disaster-specific knowledge. Although these methods are applied directly to the target data without relying on a source model trained on change detection tasks, and thus are not impacted by domain differences, their effectiveness is significantly hindered by the absence of specific knowledge. 
Secondly, the baselines (Source and Source\ding{61}) that directly utilize the source model benefit from the disaster-related knowledge in the source data, similar to our method, which aids in analyzing the target data. However, their performance declines as domain discrepancies between the source and target increases. 
For example, while the performance decrease for wildfires is minimal, the performance for other disasters is significantly affected.
Thirdly, only a few results from domain adaptation methods (TENT, SHOT, and CoTTA) surpass other baselines, despite these approaches containing mechanisms to address domain discrepancies. 
% our method consistently outperforms other baselines, but only a few results from the domain adaptation methods (i.e., TENT, SHOT, and CoTTA) surpass the baselines, despite both approaches containing mechanisms to address domain discrepancies.
Specifically, as the visual characteristics of a disaster diverge further from the source data, performance tends to decline, indicating that adapting visual style shifts based on the source model solely is not enough. \looseness=-1

\begin{table}[t!]	
% \cutcaptionup
\centering
\resizebox{0.47\textwidth}{!}	
{	
    \begin{tabular}{@{}l|llll@{}}
    \toprule
    \multicolumn{1}{c|}{\multirow{2}{*}{Component}} & \multicolumn{4}{c}{Wildfires (California, USA)} \\ \cline{2-5} 
    \multicolumn{1}{c|}{} & \multicolumn{1}{c}{Precision} & \multicolumn{1}{c}{Recall} & \multicolumn{1}{c}{F1-score} & \multicolumn{1}{c}{Accuracy} \\ \midrule
    $\squared{1}$: \ Source Only & 0.6824 & 0.8025 & 0.7376 & 0.9906 \\
    $\squared{2}$: \ $\squared{1}$ w. coarse-grained & 0.7964 & 0.6903 & 0.7396 & 0.9920 \\
    $\squared{3}$: \ Diff-SAM Only & 0.7177 & 0.3486 & 0.4693 & 0.9870 \\
    $\squared{4}$: \ $\squared{3}$ w. coarse-grained & 0.7504 & 0.3365 & 0.4647 & 0.9872 \\
    $\squared{5}$: \ Source + Diff-SAM & 0.8068 & 0.6860 & 0.7415 & \textbf{0.9921} \\
    $\squared{6}$: \ $\squared{5}$ w. coarse-grained & 0.7260 & 0.7679 & 0.7463 & 0.9914 \\
    $\squared{7}$: \ $\squared{5}$ w. Refinement & 0.7082 & 0.7819 & 0.7432 & 0.9911 \\
    $\squared{8}$: \ $\squared{6}$ w. Refinement (\pname{}) & 0.7206 & 0.7767 & \textbf{0.7476} & 0.9913 \\ \bottomrule
     \end{tabular}
}
\caption{
Ablation study result on key components.
%on fine-grained pseudo labels, including binary change maps from the source model (Source Only), segmentation model's binary confidence difference maps (Diff-SAM), and refinement process (Refinement) with/without coarse-grained pseudo labels. 
\looseness=-1
}	
\label{tab: ablation}	
\end{table}

The proposed method, addressing both visual style shifts and decision boundary shifts, consistently achieves superior results across all scenarios. This is particularly evident for disaster types not present in the source data, such as hurricanes and tsunamis, where image segmentation demonstrates highly benefits due to their distinct semantic characteristics.  %\looseness=-1 
We also confirm the robustness of our method against variations in data quantity and imbalance. %(see a descriptive analysis of the dataset in Appendix). 
\looseness=-1 \smallskip 
% \kj{Additionally, we provide a statistical analysis of the dataset in \todo{Appendix~\ref{subsec: appendix_dataset_details}}. Through this analysis, we also demonstrate the robustness of our method against variations in data quantity and imbalance.} \looseness=-1 \smallskip 

\begin{figure*}[!t]
    \centering
    \includegraphics[width=0.73\textwidth]{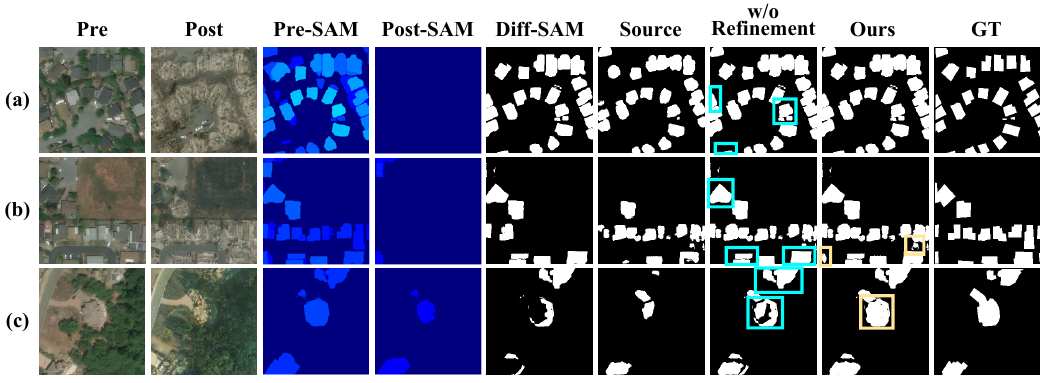}
    % \cutcaptionup
    \caption{
    Step-by-step visualization of \pname{}. From left to right, images represent pre- and post-disasters, corresponding confidence maps, confidence difference maps discretized by the optimal threshold, binary change maps from the source model, fine-grained pseudo labels without refinement, our pseudo labels, and ground-truth labels. Structures accurately identified through image segmentation appear with a cyan box, while those recognized through  refinement appear with a yellow box.
    %Additional cases on wildfires, hurricanes, and tsunamis are in \todo{Appendix}. 
    \looseness=-1
    }
    \label{fig: qualitative_analysis}
\end{figure*}

\begin{figure}[t!]

    \begin{subfigure}{0.235\textwidth}
      \includegraphics[width=\textwidth]{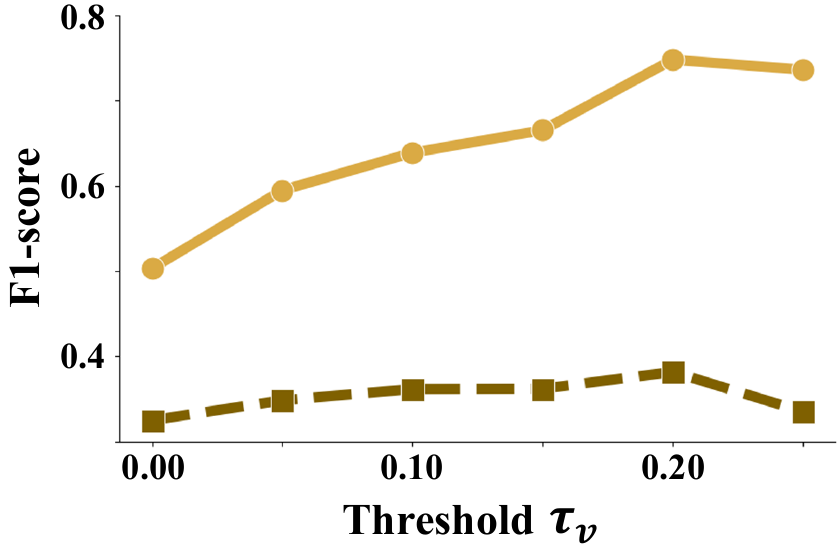}
      \caption{Wildfires}
    \end{subfigure}
    % \hfill
    \begin{subfigure}{0.23\textwidth}
      \includegraphics[width=\textwidth]{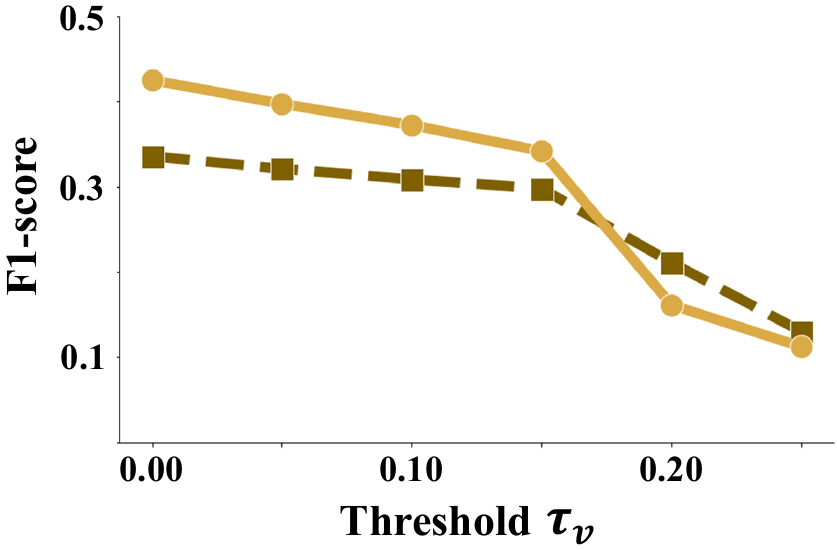}
      \caption{Hurricanes}
    \end{subfigure}
    % \cutcaptionup
    \caption{
    % Hyperparameter analysis result on the choice of $\tau_{v}$ for DAVI (solid line) and SAM (dashed line).
    Hyperparameter analysis of $\tau_{v}$. Solid line shows F1-score of \pname{}, while dashed line shows F1-score between the source model's and SAM's binary change maps across thresholds. 
    \looseness=-1}
    \label{fig: tau}
    
\end{figure}

\subsection{Component Analysis}
\label{sec: component_analysis}
\noindent\textbf{Ablation Study}. 
We validate the effectiveness of each component of the proposed method through an ablation study in Table~\ref{tab: ablation}.
The table shows the degree to which the critical components of the model.
%(Source, Diff-SAM, and Refinement) contribute to performance improvements ($\squared{1}$, $\squared{3}$, $\squared{5}$, and $\squared{8}$).
We make several observations on the results. When the domain difference between the source and target is minimal, using the source model's outputs (based on Eq. (\ref{eq: binary_cd_from_source})) is more effective than relying solely on SAM's image segmentation (based on Eq. (\ref{eq: binary_change_map_from_confidence_difference})); this comparison is evident in lines denoted in $\squared{1}$ and $\squared{3}$.  
Nevertheless, line $\squared{5}$ presents that integrating image segmentation with the source model's outputs (based on Eq. (\ref{eq: combine_binary_change_map}))  improves performance.
This finding suggests that utilizing image segmentation knowledge in unexpected regions could significantly enhance disaster assessment.
Furthermore, the results demonstrate that using coarse-grained pseudo labels mostly leads to better outcomes in terms of F1-scores, as evident in lines  ($\squared{2}$, $\squared{6}$, and $\squared{8}$) compared to cases without them ($\squared{1}$, $\squared{5}$, and $\squared{7}$).
We can confirm that masking effectively reduces false positives (FP) in the fine-grained pseudo labels.
%Additional ablation studies for the remaining target disasters, hurricanes and tsunamis, are in \todo{Appendix}.
\looseness=-1 \smallskip

\noindent\textbf{Hyperparameter Analysis}. The optimal threshold, $\tau_{v}$ in Eq. (\ref{eq: binary_change_map_from_confidence_difference}) for generating the binary change maps from the confidence difference maps produced by SAM, is reported in Figure~\ref{fig: tau}.
Our results confirm that optimal performance is achieved by selecting the threshold based on F1-score criteria, comparing binary change maps from the source model and SAM across various thresholds. 
%Further analysis related to SAM is in \todo{Appendix}.
\looseness=-1 \smallskip

\subsection{Qualitative Analysis}
\label{sec: qualitative_analysis}
In addition to the quantitative results, we also visually inspect the impact of each component within \pname{} for the case of wildfires. Figure~\ref{fig: qualitative_analysis} illustrates examples where our method benefits from the source model and SAM, refining the pseudo labels based on the target model outputs.
The cyan boxes highlight areas where SAM has positively influenced detection by identifying regions missed by the source model. The yellow boxes indicate areas improved through the refinement process, either by correcting recognized buildings (b) or identifying new structures (c).
\looseness=-1

\subsection{Real-World Case Study}
\label{subsec: case_study}
\begin{figure}[t!]
    \centering
    \includegraphics[width=0.46\textwidth]{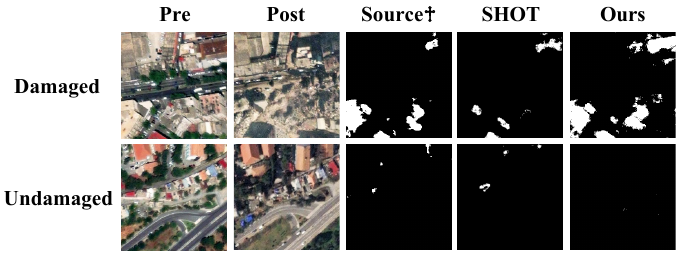}
    % \cutcaptionup
    \caption{
    Evaluation on the 2023 Türkiye earthquake. Columns show pre- and post-disaster images, predictions from baselines and our method. The top row shows damaged cases, while the bottom row shows undamaged cases. \looseness=-1
    }
    \label{fig: case_study}
\end{figure}

We present representative cases and their evaluation results using the 2023 Türkiye earthquake dataset, freely provided by Maxar, to demonstrate our method's applicability to recent real-world disasters, as shown in Figure~\ref{fig: case_study}. The dataset includes 1,648 pairs of pre- and post-disaster images.

Despite differences in geography and disaster types between the source data and the Türkiye dataset, \pname{} successfully identifies disaster-affected areas with greater accuracy than other baselines, while minimizing false identifications of undamaged areas. 
For quantitative evaluation, we conducted human annotations on a randomly selected subset of images (10\% of the total pairs). Our approach achieved an F1-score of 0.68, significantly outperforming the baselines (0.36 for Source\ding{61} and 0.42 for SHOT).
These results underscore the model’s effectiveness in rapid assessment for real-world disaster scenarios where obtaining ground-truth labels for target regions is not feasible. Application to the Türkiye scenario highlights the potential of our method to provide timely and accurate disaster response insights.
\looseness=-1

% We present representative cases and their evaluation results using the 2023 Türkiye earthquake dataset, freely provided by Maxar, to demonstrate our method's applicability to recent real-world disasters in Figure~\ref{fig: case_study}.
% This dataset includes 1,648 pairs of pre- and post-disaster images.
% Despite differences in geography and disaster types between the source data and the Türkiye dataset, \pname{} successfully identified disaster-affected areas with greater accuracy than other baselines, while minimizing false identifications of undamaged areas.
% These results underscore the model’s effectiveness in rapid assessment for real-world disaster scenario where there is no capacity to obtain ground-truth labels for target regions.
% %Additional cases are in \todo{Appendix}. 
% \looseness=-1

\section{Conclusion}
\label{sec: conclusion}
This paper introduces a test-time adaptation strategy that leverages a vision foundation model to detect fine-grained structural disaster damages. Experimental results on a real-world disaster benchmark and a newly collected disaster dataset demonstrate significant improvements in damage detection compared to other baselines including unsupervised and domain adaptation methods, particularly to unseen regions. 
To contribute to social good, we plan to make our model and implementation codes available to the research community and NGOs for humanitarian purposes. Future work will involve testing the approach in recently occurring disasters and enhancing the method by adaptively determining the extent to which the source model should be utilized, based on the differences between the source and target data. 
 \looseness=-1
% \smallskip

\section*{Acknowledgments}
This research was partly supported by the Institute for Basic Science (IBS-R029-C2). Sangyoon thanks the HKUST Institute of Emerging Market Studies with support from EY (IEMS24HS02).  \looseness=-1

\bibliography{aaai25} 

% \appendix
% \setcounter{secnumdepth}{2}
% \input{sec/appendix}

\end{document}